%% file: main.tex
\pdfoutput=1

\documentclass[11pt]{article}

\usepackage{acl}

\usepackage{times}
\usepackage{latexsym}
\usepackage{booktabs}

\usepackage[T1]{fontenc}

\usepackage[utf8]{inputenc}

\usepackage{microtype}

\usepackage{inconsolata}

\usepackage{graphicx}

\usepackage[capitalise]{cleveref}
\usepackage{subcaption}
\usepackage{xcolor}         
\usepackage{colortbl}
\usepackage{fancyvrb}
\usepackage{todonotes}
\usepackage{fontawesome5}
\newcommand{\lmda}{LMT\textsubscript{da}}

\newcommand{\itda}{INST\textsubscript{da}}

\newcommand{\snak}{\textsc{SnakModel}}
\newcommand{\llama}{{\textsc{Llama2}-7B}}
\newcommand{\llamabase}{{\textsc{Llama2}-7B\textsubscript{base}}}
\newcommand{\llamachat}{{\textsc{Llama2}-7B\textsubscript{chat}}}

\newcommand{\numresponses}{1,038}
\newcommand{\numparticipants}{63}
\newcommand{\dataset}{\textsc{DaKultur}}

\newcommand*\circled[1]{\tikz[baseline=(char.base)]{
            \node[shape=circle,draw,inner sep=.6pt] (char) {#1};}}
\newcommand{\eqcontrib}{\hspace{.3em}\raisebox{.07em}{\resizebox{1em}{!}{\circled{\tiny\faEquals}}}}

\newcommand\blfootnote[1]{%
  \begin{NoHyper}
  \begingroup
  \hspace{-1.5em}
  \renewcommand\thefootnote{}\footnote{#1}%
  \addtocounter{footnote}{-1}%
  \endgroup
  \end{NoHyper}
}

\title{\dataset{}: Evaluating the Cultural Awareness of Language Models for Danish with Native Speakers}

\author{
Max Müller-Eberstein\textsuperscript{\eqcontrib{}\faCompass\faRobot}\hspace{1em}
Mike Zhang\textsuperscript{\eqcontrib{}\faWater\faRobot}\hspace{1em} \\
\textbf{Elisa Bassignana}\textsuperscript{\faCompass\faRobot}\hspace{1em}
\textbf{Peter Brunsgaard Trolle}\textsuperscript{\faCompass}\hspace{1em}
\textbf{Rob van der Goot}\textsuperscript{\faCompass\faRobot} \\
\textsuperscript{\faCompass}IT University of Copenhagen, Denmark \hspace{1em}
\textsuperscript{\faWater}Aalborg University, Denmark  \\
\textsuperscript{\faRobot}Pioneer Center for Artificial Intelligence, Denmark \\
{\tt mamy@itu.dk}\hspace{.9em}{\tt jjz@cs.aau.dk}
}

\begin{document}
\maketitle
\blfootnote{\textsuperscript{\hspace{-1.5em}\eqcontrib{}\hspace{-.3em}} These authors contributed equally.}
\begin{abstract}
Large Language Models (LLMs) have seen widespread societal adoption. However, while they are able to interact with users in languages beyond English, they have been shown to lack cultural awareness, providing anglocentric or inappropriate responses for underrepresented language communities. To investigate this gap and disentangle linguistic versus cultural proficiency, we conduct the first cultural evaluation study for the mid-resource language of Danish, in which native speakers prompt different models to solve tasks requiring cultural awareness. Our analysis of the resulting \numresponses{} interactions from \numparticipants{} demographically diverse participants highlights open challenges to cultural adaptation: Particularly, how currently employed automatically translated data are insufficient to train or measure cultural adaptation, and how training on native-speaker data can more than double response acceptance rates. We release our study data as \dataset{}---the first native Danish cultural awareness dataset.
\footnote{Dataset and code at \url{https://mxij.me/x/dakultur}. This study was approved by the ethics committee of the IT University of Copenhagen on 24th June 2024.}
\end{abstract}

\begin{figure*}
    \centering
    \begin{subfigure}[c]{.3\textwidth}
        \centering
        \includegraphics[width=.95\textwidth]{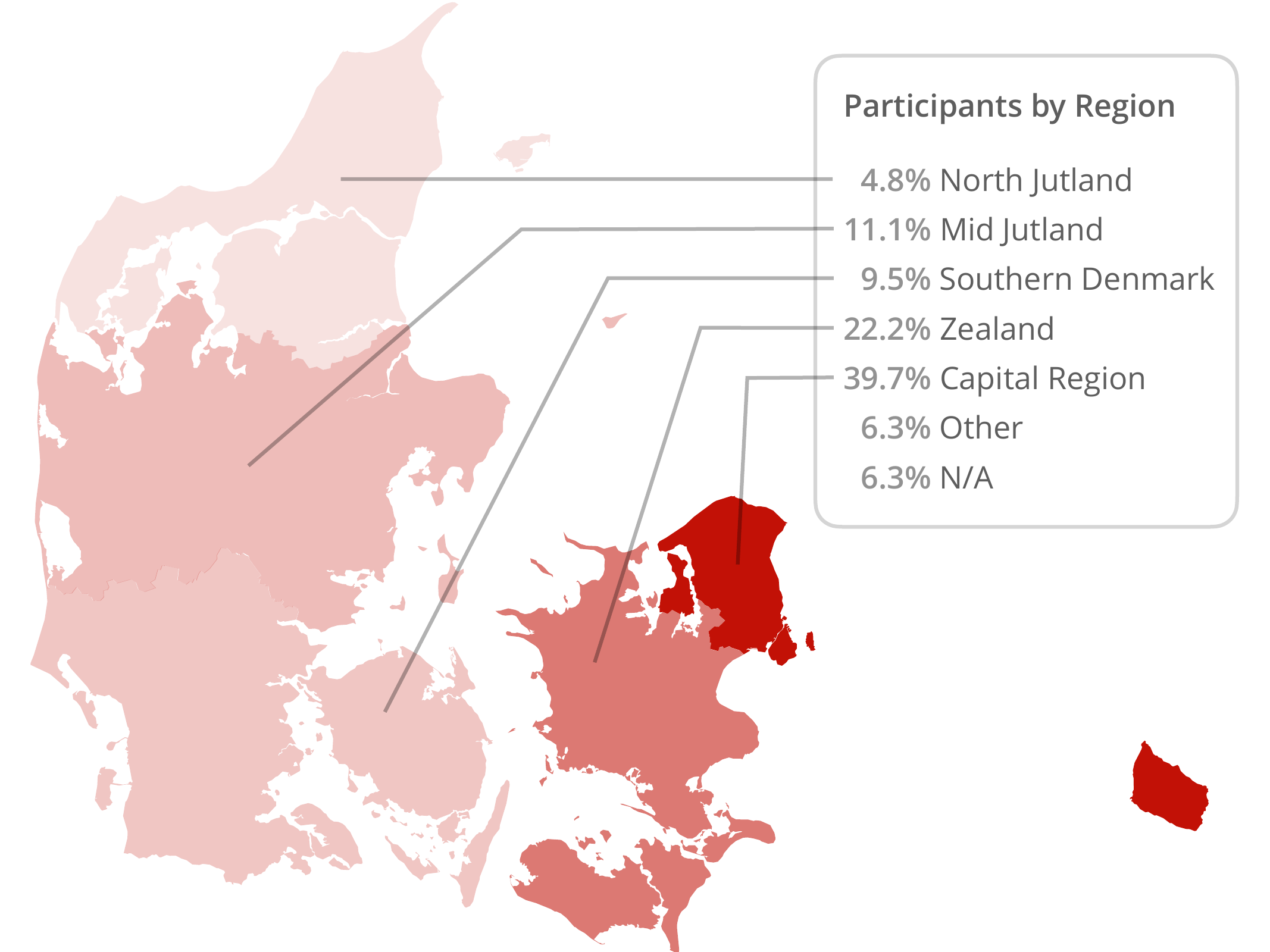}
        \caption{Participants by Region.}
        \label{fig:demographics-regions}
    \end{subfigure}
    \begin{subfigure}[c]{.3\textwidth}
        \centering
        \includegraphics[width=.95\textwidth]{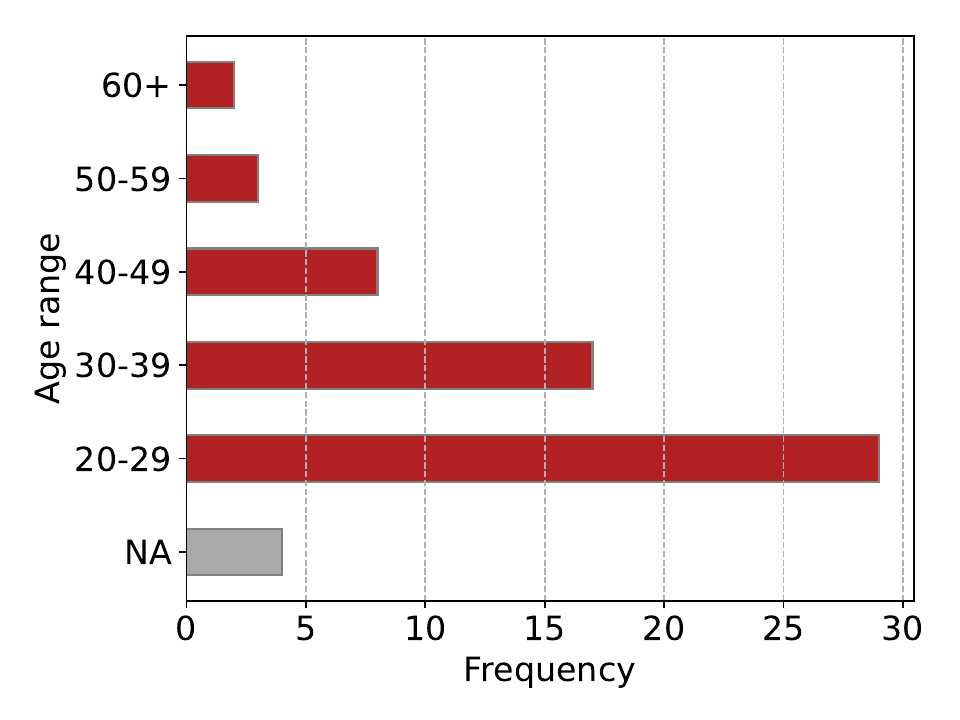}
        \caption{Participants by Age Range.}
        \label{fig:demographics-age}
    \end{subfigure}
    \begin{subfigure}[c]{.3\textwidth}
        \centering
        \includegraphics[width=.95\textwidth]{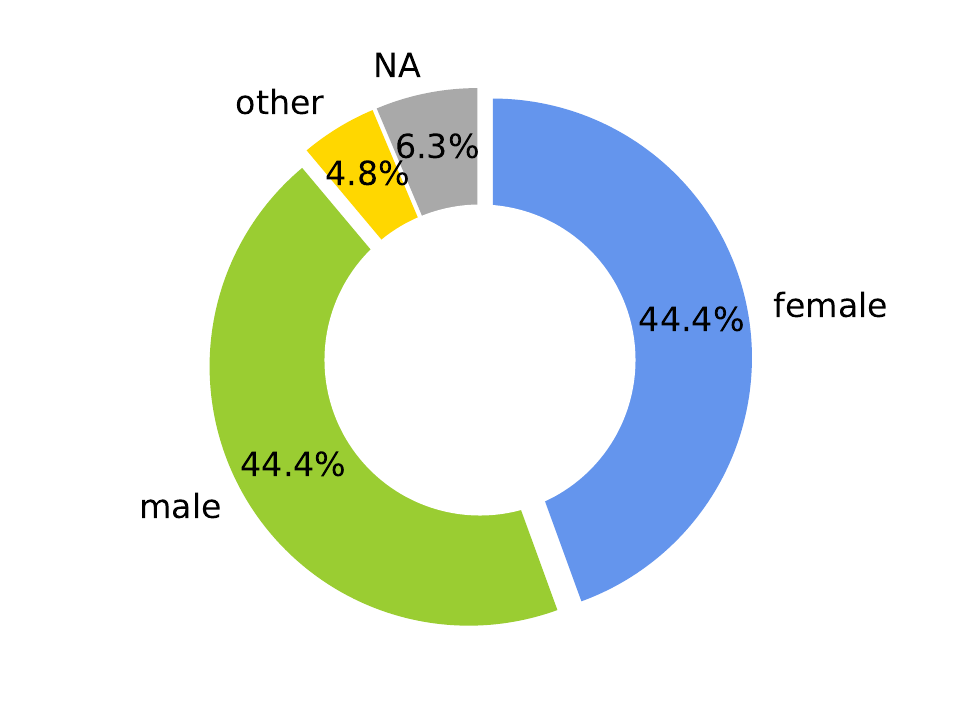}
        \caption{Participants by Gender Identity.}
        \label{fig:demographics-gender}
    \end{subfigure}
    \caption{\textbf{Demographic Statistics for Our \numparticipants{} Study Participants}, who were asked to optionally provide the region, where one grew up (\cref{fig:demographics-regions}), age range in decades (\cref{fig:demographics-age}), and gender identity (\cref{fig:demographics-gender}). 94\% of respondents opted to provide this information.}
    \label{fig:demographics}
\end{figure*}

\section{Introduction}
Culture encompasses shared beliefs, norms, and worldviews \cite{tylor1871primitive,braff2020introduction}, and tightly interweaves with language \cite{kramsch1998culture,kramsch2014aila}. These bidirectional influences affect how LLMs perform on culturally-sensitive tasks \cite{hovy-yang-2021-importance}. Contemporary LLMs are predominantly trained on English data, yet their global usage has outpaced their cultural coverage \cite{shi2023language,huang-etal-2023-languages}. Recent findings suggest that many models fail to adequately represent non-anglophone cultures \cite{hershcovich-etal-2022-challenges,zhang-etal-2023-dont,liu-etal-2024-multilingual}, resulting in culturally misaligned outputs that undermine user trust~\cite{hovy-yang-2021-importance,litschko-etal-2023-establishing,ge2024culture}.

Addressing cultural misalignment requires assessing linguistic forms, common ground, aboutness, and values \cite{hershcovich-etal-2022-challenges}. Prior efforts to improve alignment across these dimensions include the diversification of training data, as well as involving native speakers in evaluations \cite{tay-etal-2020-rather,huang-yang-2023-culturally,cao-etal-2023-assessing,naous2024beer,wang-etal-2024-countries}. However, it remains unclear which LLM adaptation strategies (i.e., data selection, training methodology) lead to the highest linguistic and cultural alignment--especially for smaller languages.

This work investigates these questions by focusing on Danish, a mid-resource language that shares typological features with English, yet differs culturally to a non-trivial degree. By performing controlled experiments with respect to linguistic and cultural adaptation, we further aim to identify guidelines for culturally adapting LLMs to languages with similar characteristics and resource constraints. Our contributions are:
\begin{itemize}
    \itemsep0em
    \item \dataset{}: The first native Danish dataset for the cultural evaluation of LLMs.
    \item A corresponding study, showing that native Danish data is critical to improving human acceptance rates (14\%$\rightarrow$42\%), and accurate automatic cultural evaluation.
    \item An analysis of the effects of demographic factors (region, age, gender) on model alignment across different cultural topics. 
\end{itemize}

\section{\dataset{}}

\subsection{Study and Data Collection Setup}
To obtain a holistic perspective on Danish culture, we construct \dataset{} based on a cultural evaluation study with native speakers in the loop. For this purpose, we build an open online interface (\cref{fig:evaluation-interface-landing}), through which we task participants to compose prompts which require an understanding of Danish culture (\cref{fig:evaluation-interface-prompting}). We then use three different language models (\cref{sec:experiment-setup}) to generate answers\footnote{Answer order was shuffled after each trial.}, which participants rate with an accept or reject, plus optional comments (\cref{fig:evaluation-interface-judgement}).

While the study is anonymous, we ask for optional demographic information (\cref{fig:evaluation-interface-demographics}), in order to assess the intra-cultural diversity of the respondents. We aim to collect only the minimal set of demographic features most likely to affect cultural responses, while not discouraging casual participation. Namely, we ask for the \textit{region} where one grew up, (the five regions of Denmark, plus \textit{other} for, e.g., people having grown up abroad), \textit{age range} in decades, and \textit{gender identity} (female, male, other).

After data collection, we manually validated the responses for relevance and correctness, and added topic annotations with a distinct set of five Danish speakers (\cref{sec:method-evaluation-human-validation}). The resulting validated study data, in the form of \dataset{}, not only serves to evaluate the cultural capabilities of the examined LLMs, but also constitutes---to the best of our knowledge---the first native Danish instruction dataset, with culturally-specific instructions, and human preference annotations.

\begin{table*}[t]
\small
  \label{tab:scandeval-results}
  \centering
  \begin{tabular}{l|cccccc|ccc}
    \toprule
    \multicolumn{1}{c|}{\textsc{Model}} & \multicolumn{6}{c|}{\textsc{Language}} & \multicolumn{3}{c}{\textsc{Culture}} \\
     & \textsc{LA} & \textsc{NER} & \textsc{SA} & \textsc{AS} & \textsc{CSR} & \textsc{QA} & \textsc{PE} & \textsc{CT} & \textbf{\textsc{DK}} \\
    \midrule
    \llamabase{}          & 33.4 & 23.7 & 61.5 & 65.5 & 29.8          & 63.5 & 38.6 & 57.7 & --- \\
    \hspace{1em} $+$ \itda{} & 36.1 & 28.5 & 62.9 & 66.4 & 29.0          & 64.4 & 49.1 & 58.5 & 13.9\\
    \llamachat{}          & 47.4 & 24.6 & 66.2 & 66.3 & \textbf{32.2} & 61.3 & 46.7 & 55.2 & --- \\
    \hspace{1em} $+$ \itda{} & 43.4 & 29.7 & 65.9 & 65.8 & 31.0          & 62.5 & 57.3 & 55.6 & 15.0 \\
    \midrule
    \rowcolor{gray!8}
    \snak{}             & \textbf{52.9} & \textbf{29.8} & \textbf{66.7} & \textbf{66.6} & 29.5 & \textbf{64.7} & \textbf{71.1} & \textbf{71.9} & \textbf{42.4} \\
    \bottomrule
  \end{tabular}
  \caption{\textbf{Results on the ScandEval Benchmark (Test) and \dataset{} (DK)}. Higher scores are better, with exact metrics depending on the sub-task (\cref{sec:experiment-setup}).  We differentiate between linguistic tasks (under \textsc{Language}), and cultural tasks (under \textsc{Culture}). Additionally, we include scores for the English \llamabase{} and \llamachat{} \cite{touvron2023llama}. The best score per sub-task is highlighted in \textbf{bold}.}
  \label{tab:results}
\end{table*}

\subsection{Study Statistics}

Our study was conducted over a period of two months, and was mainly advertised across higher educational institutions in Denmark. It attracted \numresponses{} input-response pairs with human quality judgments, from \numparticipants{} participants (after validation).

\paragraph{Demographics.} 94\% of study respondents opted to provide demographic information, for which we find coverage of all regions (\cref{fig:demographics-regions}) and gender identities (\cref{fig:demographics-gender}), as well as most age ranges except for $<$20 and $>$70 (\cref{fig:demographics-age}). We observe a slight skew towards participants who report having grown up in or around the Capital Region, that is 7\% above the expected population share, while participants from Mid/Northern Jutland and Southern Denmark are underrepresented by 4--12\%.

\begin{figure*}[t]
    \centering
    \includegraphics[width=.95\textwidth]{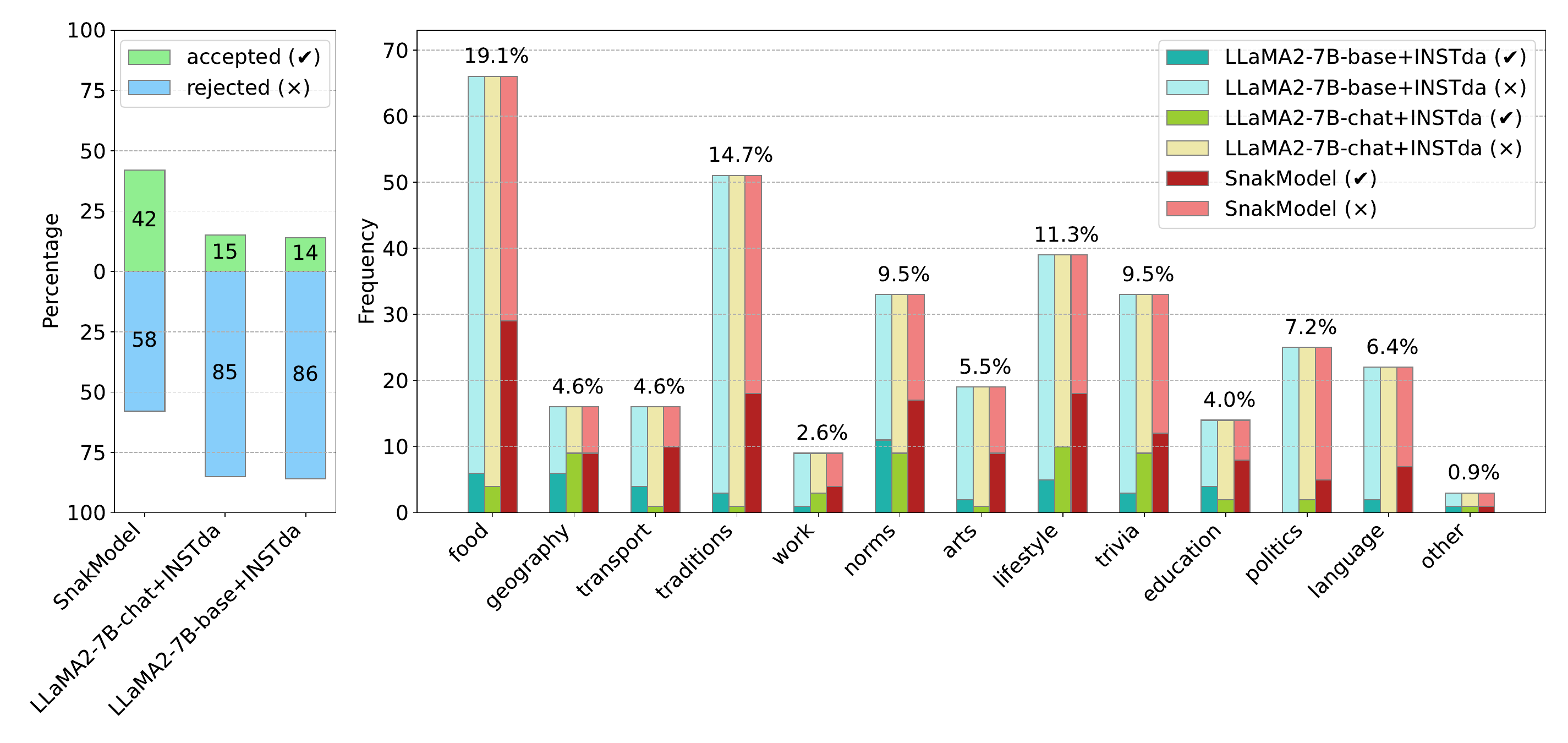}
    \caption{\textbf{Acceptance/Rejection Rates} across \snak{}, \llamachat{}+\itda{} and \llamabase{}+\itda{} as judged by participants in \dataset{}. Left: overall results; Right: results by topic.}
    \label{fig:acceptance-rate}
    \vspace{-1em}
\end{figure*}

\paragraph{Quality.} Generally, participants provided high-quality input, with 94.49\% of prompts passing our post-study validation (\cref{sec:method-evaluation-human-validation}). They further cover a diverse range of cultural concepts, as shown in the spread of topics in \cref{fig:acceptance-rate}. Prompts are more frequently phrased as questions than as instructions (e.g., ``how does a hot-dog stand look?'' versus ``describe how a hot-dog stand looks like''). Furthermore, the majority of inputs query the models' cultural knowledge directly instead of via its situational awareness of societal norms (e.g., by prompting models to resolve dilemmatic situations). As prompts in the latter format are much more time-intensive to create, this is likely to be expected. Participants further steered clear of politically and morally-charged topics, despite their anonymity. The resulting collection of cultural prompts therefore contains cultural concepts, that appear to enjoy a more uniform consensus.

\subsection{Post-study Validation}\label{sec:method-evaluation-human-validation}
Post-study, we validate and analyze the resulting data in another round of annotation with a distinct set of five Danish speakers. The study data is split across annotators, and each annotator is tasked to verify whether an input is dependent on a Danish cultural context (i.e., valid for this study), as well as which broader main topic it belongs to. For annotating topics, we employ an open coding strategy~\cite{strauss1987qualitative}, which resulted in the following 12 topics (+ other):

\begin{itemize}
    \itemsep0em
    \item \textbf{arts}: media and their place in society (e.g., ``name five popular Danish TV programs'').
    \item \textbf{education}: regarding the education system (e.g., ``which university is best to learn about AI in Denmark?'').
    \item \textbf{food}: regarding dishes and culinary traditions (e.g., ``can I serve herring on french bread?'').
    \item \textbf{geography}: regions, cities, and climate (e.g., ``where can you go on vacation in the south?'').
    \item \textbf{language}: proficiency in appropriate responses and proverbs (e.g., ``what does it mean to be a pineapple in its own juice?'').
    \item \textbf{lifestyle}: everyday activities that are not as strict as norms (e.g., ``what should I prepare when going to a Danish beach?'').
    \item \textbf{norms}: implicit rules that are followed in Danish society (e.g., ``explain the effect of `the law of Jante' on Danish mentality'').
    \item \textbf{politics}: knowledge of the political system, figures, and parties (e.g., ``how do I become a member of the regional parliament?'').
    \item \textbf{traditions}: customs and events, observed across multiple generations (e.g., ``what do you do with a 25-year-old who's single?'').
    \item \textbf{transport}: knowledge and customs regarding transportation systems (e.g., ``can you turn left on a bicycle at a traffic light?'').
    \item \textbf{trivia}: factual knowledge about people, places, historical events, sports etc., which are not part of the other topics (e.g., ``in what year was the reunification of Southern Jutland?'').
    \item \textbf{work}: procedures and behaviors, that are appropriate for a professional context (e.g., ``how do I ask my manager for a raise?'').
\end{itemize}

\section{Cultural Evaluation}\label{sec:results}

We next investigate the results of our cultural evaluation study, and compare the metrics from \dataset{} with those of automatic benchmarks.

\subsection{Experimental Setup}\label{sec:experiment-setup}

\paragraph{Models.} In our study, we use three LLMs based on \llamabase{} \citep{touvron2023llama}, adapted to Danish using distinctive training strategies: Danish language modeling training (\lmda{}), and instruction tuning on translated data (\itda{}). The corresponding models are \llamabase{}+\itda{}, \llamachat{}+\itda{}, and \snak{} (\citealp{zhang2024snakmodellessonslearnedtraining};  \llamabase{}+\lmda{}+\itda{}).

\paragraph{Automatic Evaluation.} To compare the human judgments in \dataset{} with existing automatic metrics, we employ the Danish part of ScandEval~\cite{nielsen-2023-scandeval}, across its sub-tasks on linguistic acceptability (LA from ScaLA\footnote{Based on Danish data from the Universal Dependencies dataset from \cite{kromann2004danish}.})
; named entity recognition (NER from DANSK;~\citealp{hvingelby-etal-2020-dane})
; sentiment analysis (SA from AngryTweets;~\citealp{pauli-etal-2021-danlp})
; abstractive summarization (AS from Nordjylland-News;~\citealp{kinch2023nordjylland})
; commonsense reasoning (CSR from HellaSwag;~\citealp{zellers-etal-2019-hellaswag})
; and question answering (QA from ScandiQA\footnote{Note that ScandiQA is a translation of the English MKQA dataset~\citealp{longpre2021mkqa}, and does not strictly focus on Scandinavian knowledge.}).
ScandEval further includes two culturally-oriented tasks: Danske Talemåder (PE;~\citealp{nielsen-2023-scandeval}), which prompts for meanings behind Danish proverbs, and a collection of Danish Citizenship Tests (CT;~\citealp{nielsen2024citizen}).

\subsection{Results}

\paragraph{Automatic Metrics.} Results on ScandEval (\cref{tab:results}) show that training on native Danish data (i.e., \snak{}) leads to the greatest performance gains across the board. While the unadapted English models perform comparably on some tasks, it is important to note that ScandEval employs constrained generation. When prompted without constraint, both \llamabase{} and \llamachat{} generate English responses. Instruction tuning using translated data is already sufficient to enforce Danish responses (even when prompted in English), which is why we employ the +\itda{} variants in our human study. Nonetheless, we observe that translated data is insufficient to induce much cultural knowledge into the model, as only \snak{} improves on the cultural tasks of PE and CT to a substantial degree.

\paragraph{\dataset{} Results.} In terms of acceptance rates, \snak{} obtains a rate more than twice as high compared to the other models (\cref{fig:acceptance-rate}). Nonetheless, with a maximum acceptance rate of 42\%, none of the models appears to provide particularly well-adapted responses---highlighting the gap between cultural versus linguistic adaptation. Qualitatively, we observe that answers are almost never rejected due to linguistic errors, but rather due to incorrect or incomplete factual content.

Our post-study analysis reveals that the cultural topics of \textit{food} and \textit{traditions} are most popular, and that \snak{} achieves acceptance rates over ten times as high for these topics. While training on native data improves performance across all topics, gains are larger for implicit cultural knowledge (e.g., \textit{lifestyle}, \textit{norms}) than for facts (e.g., \textit{trivia}, \textit{geography}, \textit{politics}).
In \cref{app:demographic_results}, we further show how topics and acceptance rates vary by demographics. Female-identity participants tend toward \textit{food}, \textit{lifestyle}, \textit{education}, and \textit{norms}, while male-identity participants focus more on \textit{politics}, \textit{trivia}, and \textit{geography}. Additionally, younger participants and those from the capital region report slightly higher acceptance rates.

\section{Conclusion}
In this work, we introduced \dataset{}---the first native Danish cultural evaluation dataset. By constructing it via a native-speaker-driven evaluation study, and applying a thorough post-study validation, we are able to share 1,038 high-quality input-response pairs for future Danish NLP research. Our cultural evaluation using \dataset{} highlights that language modeling training using native data is already sufficient to more than double human-judged cultural awareness---especially for popular cultural topics. Simultaneously, the maximum acceptance rate of 42\% highlights that more research is needed to fully align anglocentric LLMs to smaller language communities, such as Danish. In terms of evaluation methodologies, the fact that human judgments align more with the smaller, yet culturally-relevant and non-translated sub-tasks of the automatic ScandEval benchmark (PE, CT, as well as LA) is encouraging, since small amounts of high-quality data may already be sufficient to accurately estimate an LLM's cultural awareness.

\clearpage
\clearpage

\section*{Limitations}

While we strive for broad coverage of the Danish cultural landscape, culture itself has a high degree of inherent subjectivity and variability. As such, future work using \dataset{} should be cognizant of the context in which its data was obtained. Our cultural evaluation study was advertised primarily at higher educational institutions. Although we are aware of word-to-mouth advertisement stretching to demographic groups beyond this initial cluster (as evidenced by the range of represented age groups), the study likely does not capture the full breadth of the Danish cultural landscape. By gathering demographics for intra-cultural differences with regard to topics and user acceptance rates, we nonetheless aim to enable analyses with respect to how much cultural consensus might vary with respect to different topics. We believe this is crucial information for practitioners designing downstream systems, as contemporary models seem to, for instance, align slightly better with male-identity participants under 30 from the capital region.

On the technical side, we hope that future work will be able to validate our findings across more base models and languages. Our choice of Danish and \llama{}-based models was primarily driven by data and compute resource availability. Similarly, while \dataset{} can theoretically be used for small-scale instruction tuning or model alignment, its size is far from contemporary, automatically generated datasets. For cultural evaluation purposes, we nonetheless believe that it offers a representative out-of-the-box solution for developers of future Danish LMs.

\section*{Acknowledgments}
We thank the NLPnorth group at ITU and the AAU-NLP group at AAU for feedback on earlier version of this draft. 
Elisa Bassignana is supported by a research grant (VIL59826) from VILLUM FONDEN. Mike Zhang is supported by a research grant (VIL57392) from VILLUM FONDEN.

\bibliography{custom}

\appendix

\include{appendix}

\end{document}

%% file: appendix.tex
\section*{Appendix}

\begin{figure*}[t]
    \centering
    \begin{subfigure}[c]{.24\textwidth}
        \centering
        \includegraphics[width=.9\textwidth]{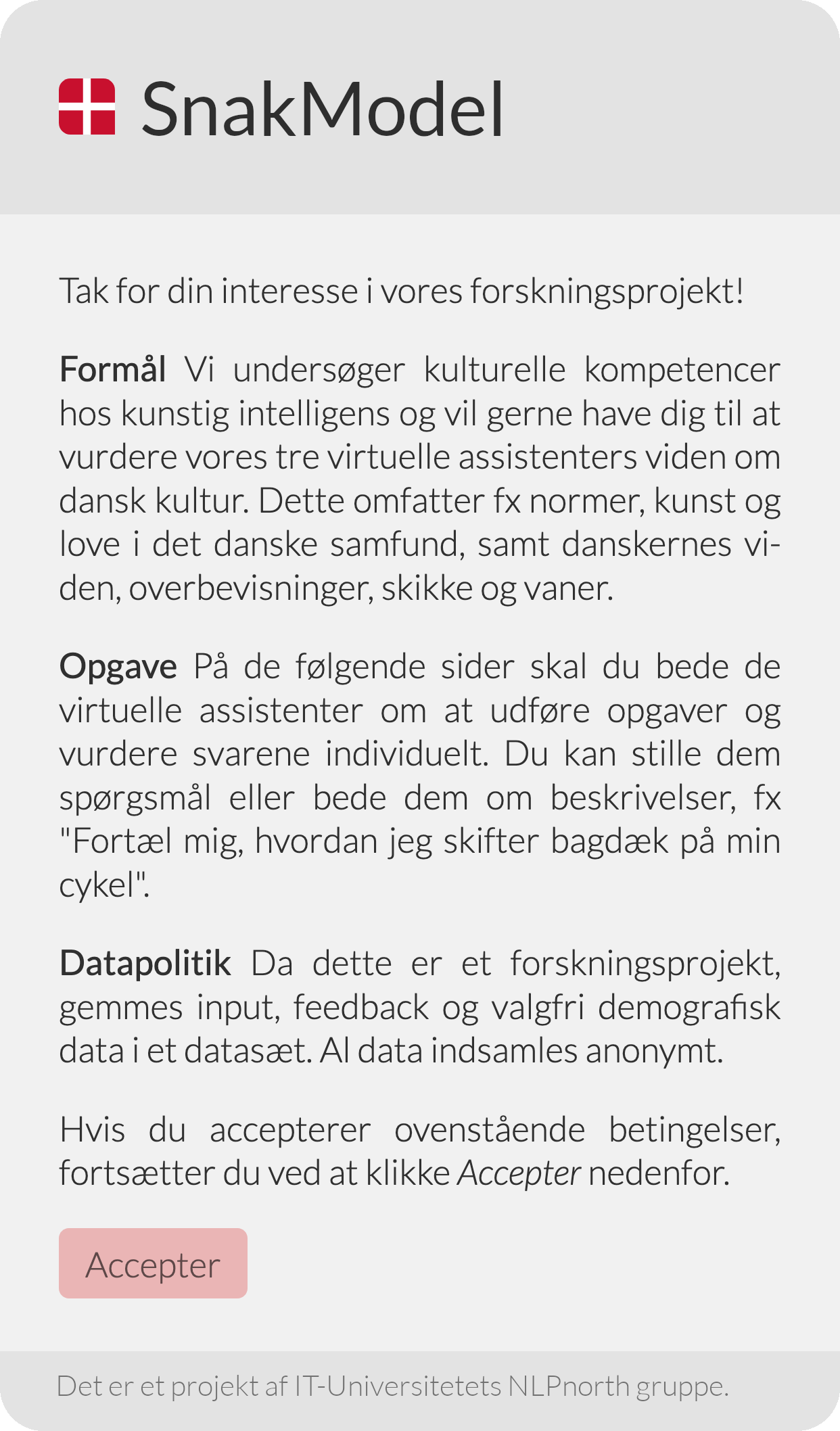}
        \caption{Guidelines.}
        \label{fig:evaluation-interface-landing}
    \end{subfigure}
    \begin{subfigure}[c]{.24\textwidth}
        \centering
        \includegraphics[width=.9\textwidth]{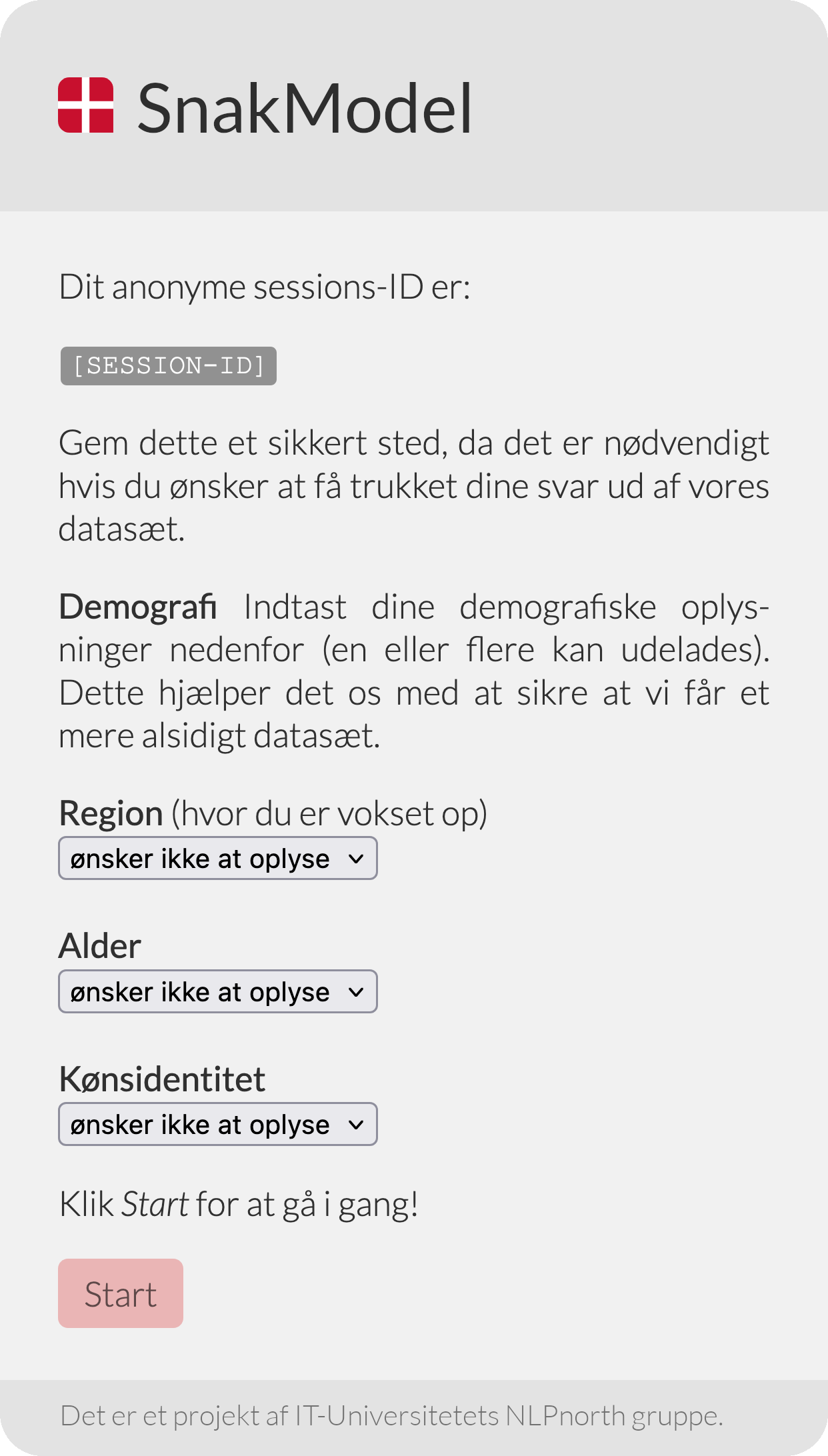}
        \caption{Demographics.}
        \label{fig:evaluation-interface-demographics}
    \end{subfigure}
    \begin{subfigure}[c]{.24\textwidth}
        \centering
        \includegraphics[width=.9\textwidth]{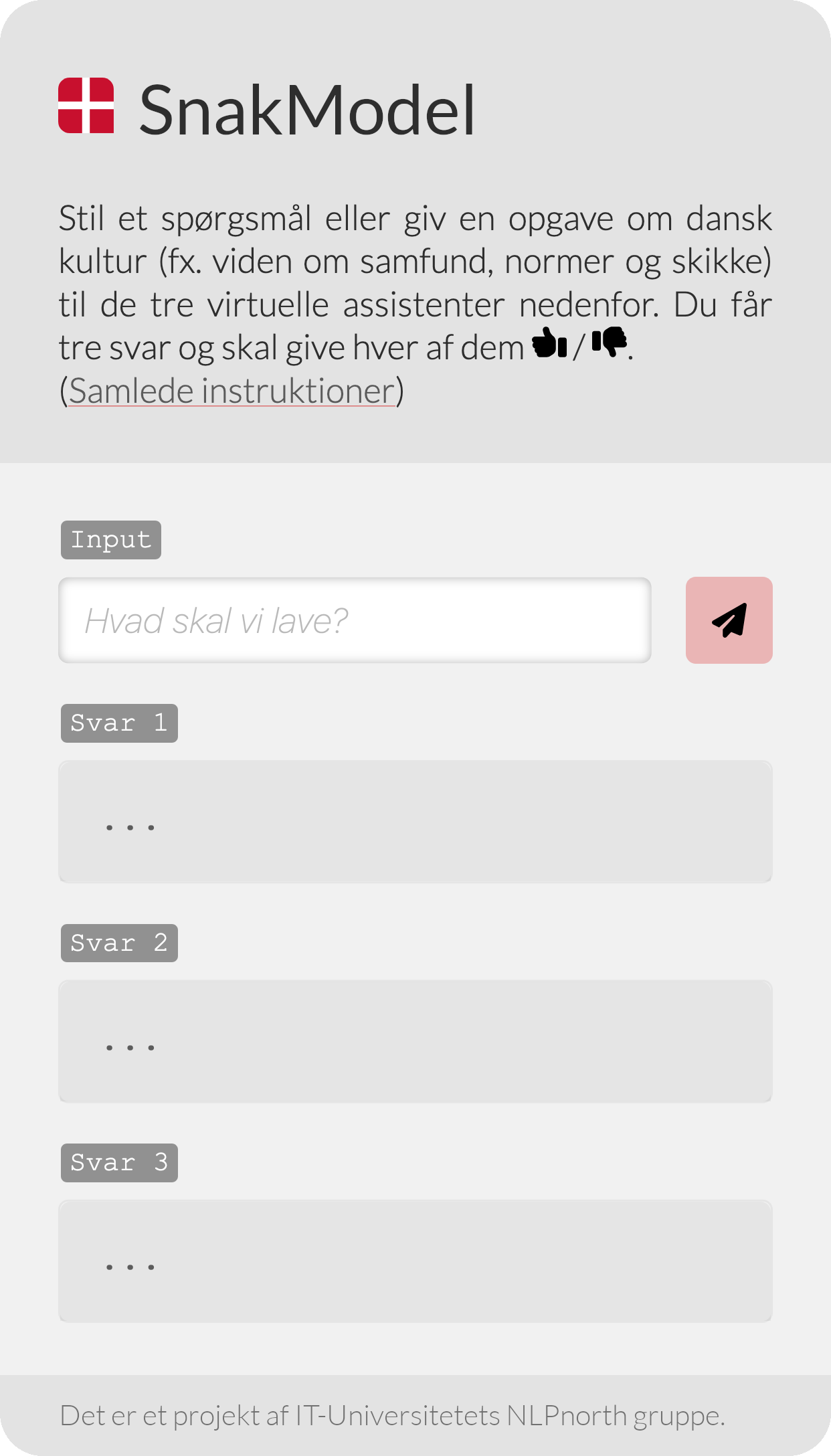}
        \caption{Prompt Interface.}
        \label{fig:evaluation-interface-prompting}
    \end{subfigure}
    \begin{subfigure}[c]{.24\textwidth}
        \centering
        \includegraphics[width=.9\textwidth]{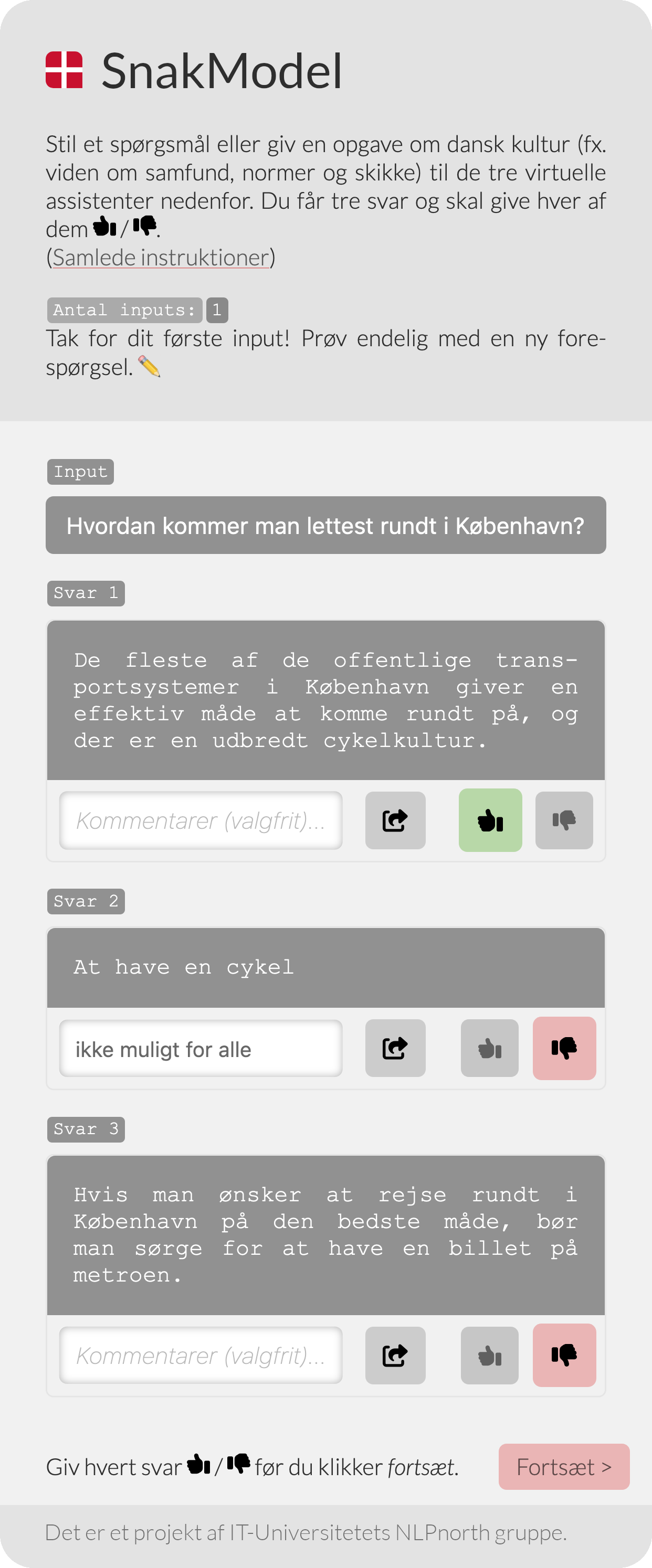}
        \caption{Evaluation Interface.}
        \label{fig:evaluation-interface-judgement}
    \end{subfigure}
    \caption{\textbf{Study Interface for Human Cultural Evaluation}. Participants are guided through the guidelines (\cref{fig:evaluation-interface-landing}), optional demographic registration (\cref{fig:evaluation-interface-demographics}), before being asked to prompt the three LLMs simultaneously (\cref{fig:evaluation-interface-prompting}), and to evaluate the model responses (\cref{fig:evaluation-interface-judgement}). Translations of the guidelines, interface, and examples can be found in \cref{app:translation}.}
    \label{fig:evaluation-interface}
\end{figure*}

\section{Study Interface}
We build a web-based evaluation interface (study flow shown in \cref{fig:evaluation-interface}), which allows study participants to prompt the three LLMs simultaneously\footnote{Note that the order in which responses are displayed is randomized with each prompt.} with tasks and questions, that require cultural awareness (\cref{fig:evaluation-interface-prompting}), and to rate (accept/reject) and comment on the models' responses (\cref{fig:evaluation-interface-judgement}). The study guidelines (\cref{fig:evaluation-interface-landing,fig:evaluation-interface-prompting}) broadly lay out which dimensions of cultural awareness the study aims to investigate---i.e., common ground, aboutness, objectives and values, in addition to linguistic form and style \cite{hershcovich-etal-2022-challenges}, which is implicit, given the study's monolingual nature. Following prior work on culturally diverse dataset creation \cite{liu-etal-2021-visually}, we opted for an elicitation setup in order to avoid biasing responses towards a limited set of cultural concepts and topics.

While the study is conducted anonymously, we ask for optional demographic information (\cref{fig:evaluation-interface-demographics}), in order to assess the intra-cultural diversity of the respondents. For this purpose, we aimed to collect only the minimal set of demographic features, that we deemed most likely to affect cultural responses, while not discouraging casual participation. Namely, we ask for the region, where one grew up, (the five regions of Denmark, plus \textit{other} for, e.g., people having grown up abroad), age range in decades, and gender identity (female, male, other).

In test trials, we noticed that, while participants intuitively prompted for a wide variety of culturally-relevant topics, they typically did so in a multi-turn conversational manner, which our single-turn, instruction-tuned models often fail to answer. For instance, the prompt ``Hello! Could you tell me about [...]?'', frequently produces the response, ``Yes, I can.'', with no further relevant content. To encourage single-turn instruction-style inputs, we iterated over multiple guideline formulations, of which we found, ``Ask one question or give one task about Danish culture [...] to the three virtual assistants below'', to produce the most compatible results (see full translations in \cref{app:translation}).

\section{Translations}\label{app:translation}

\subsection{Landing Page with Guidelines}

Thanks for your interest in our research project! \\
\\
\textbf{Purpose} We examine cultural skills/competencies with artificial intelligence and would like you to assess our three virtual assistants' knowledge of Danish culture. This includes, for example, norms, art and laws in Danish society, as well as Danes' knowledge, beliefs, customs and habits. \\
\\
\textbf{Task} On the following pages, you should ask the virtual assistants to perform tasks and assess their answers one-by-one. You can ask them questions or ask them for descriptions, e.g., ``tell me how to change the back tire of my bike''. \\
\\
\textbf{Data policy} As this is a research project, input, feedback and optional demographic data are stored in a dataset. All data is collected anonymously. \\
\\
If you agree to the above terms, continue by clicking Accept below. \\

\subsection{Demographic Information}

Your anonymous session ID is: \\
\texttt{SESSION\_ID}\\
\\
Save it in a safe place since it is required if you would like to get your answers removed from our dataset.\\ 
\\
\textbf{Demographics} Enter your demographic information below (one or more can be omitted). This helps us to ensure that we get a more diverse data set.\\
\\
\textbf{Region} (where you grew up)\\
\textit{do not wish to disclose}\\
\textbf{Age}\\
\textit{do not wish to disclose}\\
\textbf{Gender Identity}\\
\textit{do not wish to disclose}\\
\\
Click \textit{Start} to get started!\\

\subsection{Prompt Interface}

Ask one question or give one task about Danish culture (e.g., knowledge of society, norms and customs) to the three virtual assistants below. You will receive three answers, which you can each rate with a thumbs-up/down.\\
\\
\textbf{Input}\\
\textit{What shall we do?}\\
\\
\textbf{Answer 1}\\
\textbf{Answer 2}\\
\textbf{Answer 3}\\

\subsection{Response Evaluation Interface}\label{app:translation-evaluation-interface}

Thanks for your first input! Go ahead, and try another request! \\
\\
\textbf{Input}\\
\textit{What's the easiest way to get around in Copenhagen?}\\
\\
\textbf{Answer 1}\\
\texttt{Most of the public transport systems in Copenhagen provide an effective way to get around, and there is a widespread cycling-culture.}\\
\\
\textbf{Answer 2}\\
\texttt{To have a bike}\\
\textit{Comment: not possible for everyone}\\
\\
\textbf{Answer 3}\\
\texttt{If you want to travel around Copenhagen, you should make sure to have a ticket for the subway.}\\
\\
Give each answer a thumbs-up/down before clicking \textit{Continue}. \\

\section{Topics and Acceptance Rates per Demographic}
\label{app:demographic_results}
\begin{figure*}[!ht]
    \includegraphics[width=0.9\linewidth]{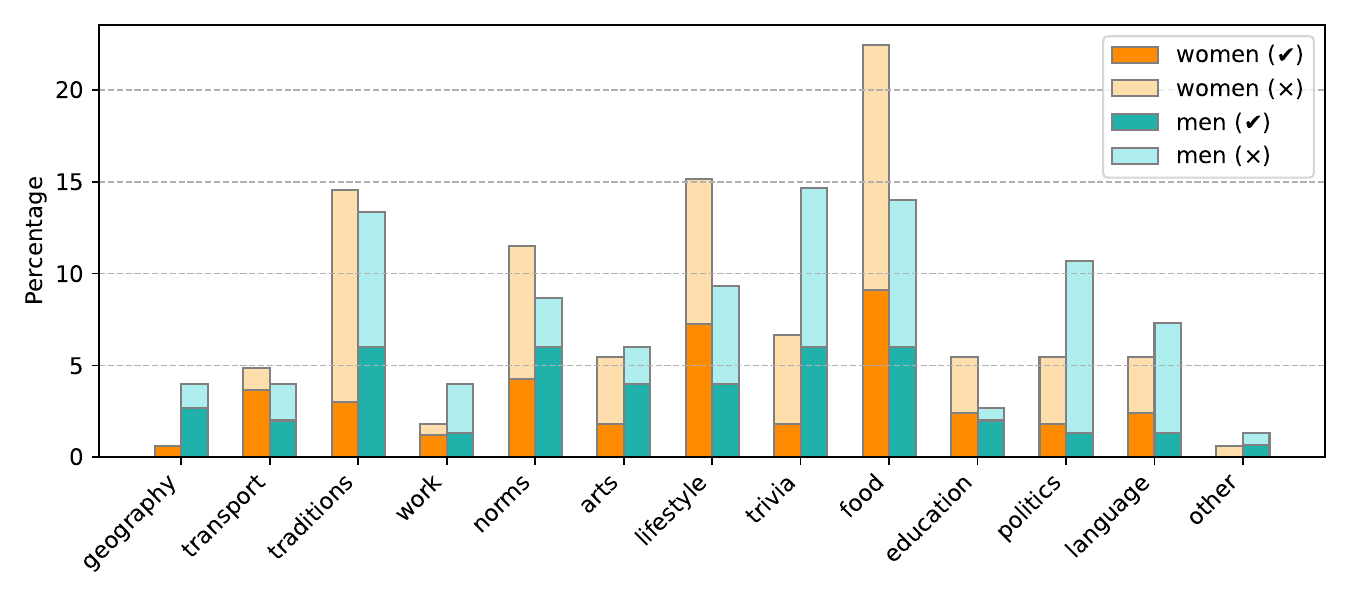}
    \caption{\textbf{Acceptance/Rejection Rates and Distribution across Topics} for the female/male gender identity demographic groups.}
    \label{fig:topic-gender}    
\end{figure*}

\begin{figure*}
    \includegraphics[width=0.9\linewidth]{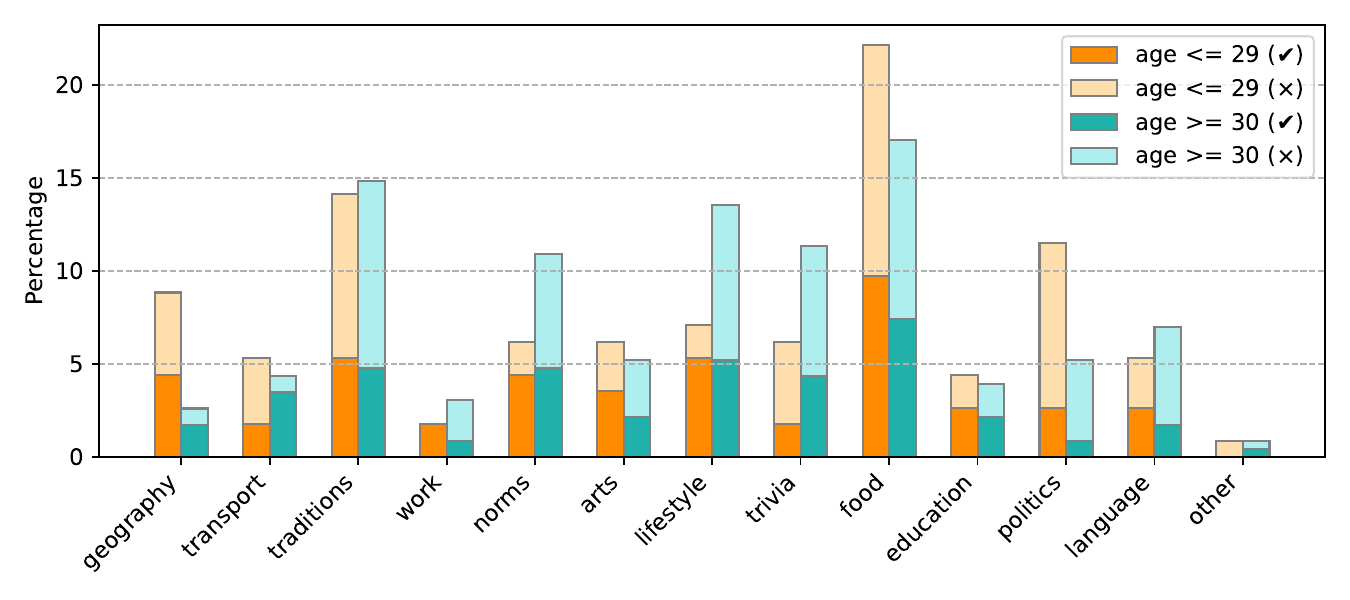}
    \caption{\textbf{Acceptance/Rejection Rates and Distribution across Topics} for the age ranges $>=$ 29 and $<=$ 30.}
    \label{fig:topic-age}    
\end{figure*}

\begin{figure*}
    \includegraphics[width=0.9\linewidth]{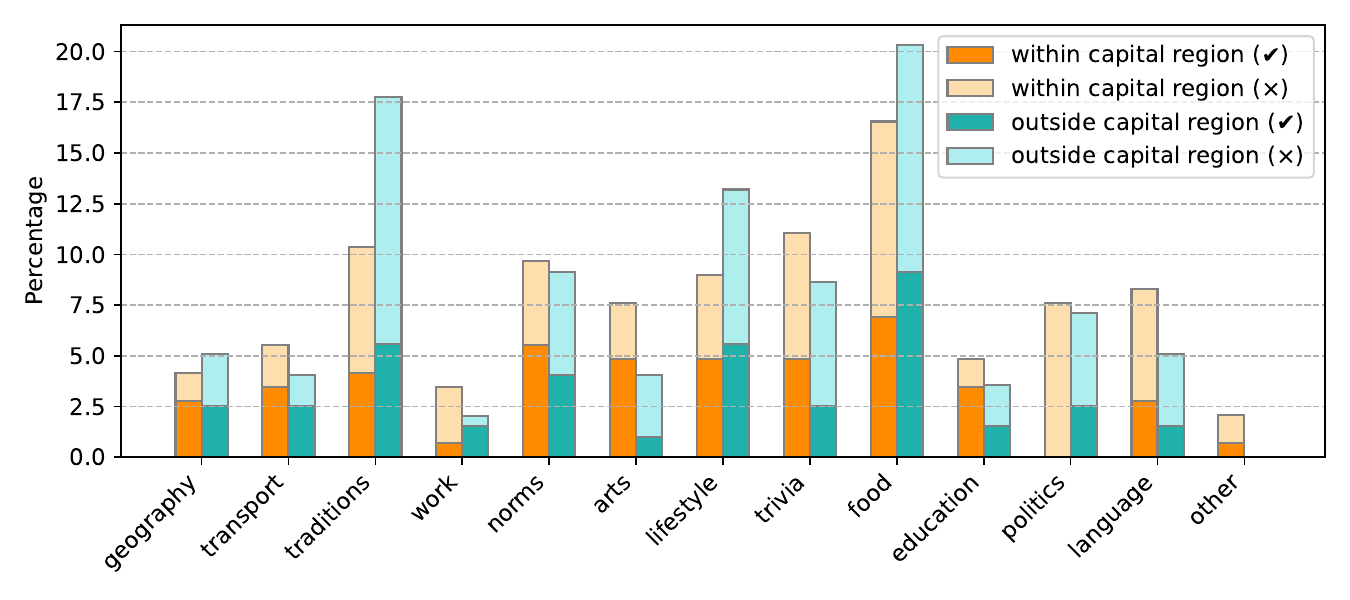}
    \caption{\textbf{Acceptance/Rejection Rates and Distribution across Topics} for participants from the capital versus other regions.}
    \label{fig:topic-region}    
\end{figure*}

For each demographic dimension, we merge the available categories into two groups, in order to have a large enough amount of information to compare. This leads to splits along male/female, $<30$/$\geq30$, capital region/other regions. The distribution as well as acceptance rates are shown in \cref{fig:topic-gender} for gender, \cref{fig:topic-age} for age, and \cref{fig:topic-region} for region.

